\documentclass{article}

\usepackage{arxiv}

\usepackage[utf8]{inputenc} % allow utf-8 input
\usepackage[T1]{fontenc}    % use 8-bit T1 fonts
\usepackage{hyperref}       % hyperlinks
\usepackage{url}            % simple URL typesetting
\usepackage{booktabs}       % professional-quality tables
\usepackage{amsfonts}       % blackboard math symbols
\usepackage{nicefrac}       % compact symbols for 1/2, etc.
\usepackage{microtype}      % microtypography
\usepackage{lipsum}
\usepackage{amsmath}
\usepackage{algorithm,algorithmic}

\usepackage{graphicx}
\usepackage{arydshln}
\usepackage{xcolor}
\usepackage{bbm}
\title{ Precision-Recall Curve (PRC) Classification Trees}

\author{
  Jiaju Miao\thanks{Corresponding author. \emph{Tel.}: +1(631)-913-5273. \emph{Website}: https://sites.google.com/stonybrook.edu/jiajumiao.}  \\ Department of Applied Mathematics and Statistics\\
 Stony Brook University\\
  Stony Brook, NY 11794 \\
  \texttt{jiaju.miao@stonybrook.edu} \\
   \And
 Wei Zhu\\ Department of Applied Mathematics and Statistics\\
  Stony Brook University\\
  Stony Brook, NY 11794 \\
  \texttt{wei.zhu@stonybrook.edu} \\
}

\begin{document}
\maketitle

\begin{abstract}
 The classification of imbalanced data has presented a significant challenge for most well-known classification algorithms that were often designed for data with relatively balanced class distributions. Nevertheless skewed class distribution is a common feature in real world problems. It is especially prevalent in certain application domains with great need for machine learning and better predictive analysis such as disease diagnosis, fraud detection, bankruptcy prediction, and suspect identification. In this paper, we propose a novel tree-based algorithm based on the area under the precision-recall curve (AUPRC) for variable selection in the classification context. Our algorithm, named as the “Precision-Recall Curve classification tree”, or simply the “PRC classification tree” modifies two crucial stages in tree building. The first stage is to maximize the area under the precision-recall curve in node variable selection. The second stage is to maximize the harmonic mean of recall and precision (F-measure) for threshold selection. We found the proposed PRC classification tree, and its subsequent extension, the PRC random forest, work well especially for class-imbalanced data sets. We have demonstrated that our methods outperform their classic counterparts, the usual CART and random forest for both synthetic and real data. Furthermore, the ROC classification tree proposed by our group previously has shown good performance in imbalanced data. The combination of them, the PRC-ROC tree, also shows great promise in identifying the minority class.
\end{abstract}

% keywords can be removed
\keywords{Precision-Recall curve \and Imbalanced data \and PRC random forest \and PRC-ROC random forest}

\section{Introduction}
\label{sec:Intro}
 Knowledge Discovery and Data Mining (KDD) is an interdisciplinary area focusing on methodologies for extracting useful information from data. Classification is a fundamental tool in this domain. In the terminology of machine learning\cite{Ref1}, classification is considered an instance of supervised learning, i.e., learning where a training set of correctly identified observations is available. The task of the constructed classifier is to predict class labels given any input information such as financial crisis prediction\cite{Ref2}\cite{Ref3}. In this introductory section, we provide a brief overview of some well-developed classification algorithms to shed light on the complex issues of class imbalance.\\
       Decision trees can be used to visually and explicitly represent decisions and the decision making process. A typical tree consists of three components. The inner tree nodes represent  features with thresholds leading to the branching of edges eventually ended up in terminal nodes called leaves labeled with class decisions. In tree building, the decision tree model is built by recursively splitting the training data set based on a locally optimal criterion (typically information gain or gini impurity until certain stopping criteria are met. After building the decision tree, a tree pruning step is performed to prevent overfitting by reducing the size of the tree. In the presence of the class imbalance, more branches may need to be built in order to identify the minority class. CART is a well-known tree algorithm\cite{Ref4}.\\
       A random forest (RF) or random decision forest\cite{Ref5} is an ensemble learning method obtained by generating many decision trees and then combine their decisions. RF is a popular bagging algorithm which incorporates those trees\cite{Ref6}. Two types of randomness are built in the process. First, tree bagging is employed to select a bootstrap sample from original the original data for each decision tree. The bootstrap resampling method was introduced by Bradley Efrons\cite{Ref7}. It is implemented by sampling with replacement from an existing sample data independently with same size and conducting experiments among them. Second, a subset of features is randomly selected to generate the best split in each tree building process. The reason in this is to decrease the correlation between trees and reduce overfitting problem. \\ 
       A Support Vector Machine (SVM) introduced by Cortes is a discriminative classifier based on a separating hyperplane\cite{Ref8}.  Given training data, for two-group (i.e., bi-class) classifications, the algorithm outputs an optimal hyperplane where the margin is maximal. Akbani\cite{Ref9} illustrates that SVMs can be ineffective in determining the class boundary when the distribution is skewed.\\
       A neural network is a series of algorithms that aims to discover the underlying relationships in a set of training data through a process that mimics the way the human brain operates. It is a beautiful bio-inspired programming paradigm. However, some experimental results demonstrate that its performance when dealing with imbalanced data is not satisfactory. The principal reason is the minority class cannot be adequately weighted\cite{Ref10}.\\
       Besides these arguably the most fundamental classifiers, there are still many other classification algorithms which we will not be able to introduce exhaustively here. Our goal is modest as we aim to develop a better decision tree algorithm, and on top of which, a better random forest algorithm, and them compare them to these four original and fundamental algorithms: CART, Random Forest (RF), SVM, and Neural Network, plus the weighted version of CART and RF that in a way reflect the current state-of-the-art on combing class-imbalance. In the next section, we shall introduce several common solutions to the class-imbalance problem.

\section{Class imbalance problem}
\label{sec:2}
      A number of solutions to the class imbalance problem have been presented in the literature. Basically they can be divided into 3 broad categories, i.e., cost-sensitive learning, data level and algorithm level approaches.
\subsection{Cost-sensitive learning}
\label{sec:2.1}
       Cost-sensitive learning first introduced by Elkan\cite{Ref11} has shown to be an effective technique for incorporating the different misclassification costs into the classification process. The goal of this type of learning is to minimize the total cost of misclassification. The key point is to assign different values to the false positives and negatives. If we focus on identifying the minority which is the common case, the cost for labeling a positive example as negative can be higher than the cost for labeling a negative example as positive. Weighted Random Forest is the extended version that makes random forest more suitable for learning from extremely imbalanced data based on the idea of cost sensitive learning\cite{Ref12}\cite{Ref13}. Kim\cite{Ref14} proposed the weighted support vector machine which imposes weights to the loss term. Likewise, some other learning algorithms have their cost-sensitive correspondings, such as cost-sensitive decision tree, cost-sensitive neural network\cite{Ref15}\cite{Ref16}\cite{Ref17}\cite{Ref18}\cite{Ref19}, etc.
\subsection{Data level approaches}
\label{sec:2.2}
       The data-level approaches can also be interpreted as resampling techniques. That is, we can either undersample the majority class or oversample the minority class to attain the balanced distribution (the ratio of class distribution is 1:1).\\
    Undersampling can be defined as removing some observations of the majority class. Some algorithms have been applied to alleviate the skewness, e.g., random undersampling and cluster-based undersampling\cite{Ref20}. However an obvious drawback is that we are removing information that may be valuable which can then lead to an underfitting problem.\\
      Similarly, oversampling before splitting can achieve more minority observations. Well-known algorithms such as SMOTE\cite{Ref21} and ADASYN\cite{Ref22} have been widely implemented in dealing with imbalanced classification problems. However, it can cause overfitting and poor generalization.
\subsection{Algorithm level approaches}
\label{sec:2.3}
It is always a good rule of thumb to modify the training procedure of algorithms to boost its performance with uneven datasets. To develop an algorithmic solution, one needs in-depth knowledge of both the classifier learning algorithm and the application domain. Based on the receiver operating characteristics (ROC), our group had invented the ROC classification trees and the ROC random forest to identify the rare cases. Song suggested choosing the threshold with the largest harmonic mean of sensitivity and specificity as the node splitting criterion\cite{Ref23}\cite{Ref24}. Yan\cite{Ref25} extended the bi-class classification ROC Tree and ROC RF\cite{Ref23} to multiple classes. Our proposed algorithm PRC classification trees, and later the PRC random forest, are developed from a different perspective as will be introduced in details in the following chapter.
\subsection{Ensemble methods}
\label{sec:2.4}
       It is well known that through combining classifiers one can improve the prediction accuracy. It is effective to embed the data level approaches in boosting procedure which is one of the most popular combination techniques. Sun\cite{Ref26} introduced the cost items into the learning framework of AdaBoost for improving the classification of imbalanced data\cite{Ref27}. Boosting has been proven to be a general and effective method for improving the accuracy of a given learning algorithm\cite{Ref28}\cite{Ref29}. Nitesh\cite{Ref30} combined the SMOTE algorithm and the boosting procedure and WOTBoost\cite{Ref31} adjusted its oversampling strategy at each round of boosting to synthesize more targeted minority data samples. Similarly, RUSBoost alleviated class imbalance by introducing random undersampling techniques into the standard boosting procedure\cite{Ref32}. MEBoost mixed two different weak learners with boosting to improve the performance on imbalanced datasets\cite{Ref33}. So far, many learning methods have had their boosting counter parts, such as DataBoost\cite{Ref34}, EUSBoost\cite{Ref35}.

\subsection{Performance evaluation}
\label{sec:2.5}
Performance metric plays a key role in evaluating the classifier. Additionally, it is a guideline for the modeling process of classifiers. Accuracy is a traditional measure to evaluate the performance. It may be misleading, however, for a data set with unbalanced class distribution. Suppose we give every observation the prevalent class label, the accuracy can be high but the performance for the rare class is very poor. The following is a brief summary on some commonly used performance metrics.
\subsubsection{Confusion matrix}
\label{sec:2.5.1}
In the bi-class scenario, positive and negative class samples can be categorized into four groups based on the classification process. This can be denoted as the confusion matrix shown in Tabel 1.
\begin{table}
\centering
\caption{Confusion matrix}
\label{tab:1}       % Give a unique label
\begin{tabular}{lll}
\hline\noalign{\smallskip}
& Predicted as Positive & Predicted as Negative  \\
\noalign{\smallskip}\hline\noalign{\smallskip}
Actually Positive & True Positive (TP) &False Negative (FN) \\
Actually Negative &  False Positive (FP) & True Negative (TN) \\
\noalign{\smallskip}\hline
\end{tabular}
\end{table}
\\Several measures can be derived from this matrix:
\begin{align*}
  &True \, positive\, rate (Recall/Sensitivity): TPR=\frac{TP}{TP+FN}\tag{1}\\
  &False \, negative\, rate : FNR=\frac{FN}{TP+FN}\tag{2}\\
  &True\, negative\, rate (Specificity):TNR=\frac{TN}{TN+FP}\tag{3}\\
  &False\, positive\, rate :FPR=\frac{FP}{TN+FP}\tag{4}\\
  &Positive\, predictive\, value (Precision):PPV=\frac{TP}{TP+FP}\tag{5}\\
  &Negative\, predictive\, value :NPV=\frac{TN}{TN+FN}.\tag{6}\\
\end{align*}
\subsubsection{F-measure}
\label{sec:2.5.2}
For situation where the performance of positive class is the priority, two measures of the indicators described earlier are crucial – the true positive rate and the positive predictive value. Specifically, the true positive rate is defined as recall referring to the percentage of positive class samples that are correctly classified. Correspondingly, the positive predictive value is defined as precision denoting the percentage of predictive positive class samples that are correctly classified. \\
      F-measure is devised to integrate these two measures. In principle, it is the harmonic mean of precision and recall:
      \begin{equation}
F_{measure}=\frac{2}{recall^{-1}+precision^{-1}} .\tag{7}
\end{equation}
      \\In general, F-measure tends to be closer to the smaller one of the measures. Additionally, a high value indicates that both two measures are reasonably high and convincing.

\section{Methodology }
\label{sec:3}
\subsection{ROC Curve and the area under the ROC curve (AUC)}
\label{sec:3.1}
ROC curve is a performance measurement for classification problem at various thresholds settings. The curve is plotted based on the ROC space with True positive rate (TPR) against the false positive rate (FPR) where TPR is on y-axis and FPR is on the x-axis. In the ROC space, a ROC curve is created by connecting all the pairs of TPR and FPR at each threshold for a specific classifier. It is a common method to calculate the area under the ROC curve (AUC) to compare the measurement of separability among different classifiers. The higher AUC refers to a stronger ability of distinguishing between classes.  There are several methods to perform a computational analysis of AUC estimators. The way we used in our analysis is to apply the trapezoidal estimators.  The estimation is based on an approximation of the entire area by summing up all subareas and the formula is:
\begin{equation}
   AUC_{trapezoid}=\frac{1}{2}\cdot\sum_{i=1}^{n}(f_{i+1}-f_{i})\cdot(t_{i+1}+t_{i})\tag{8}
\end{equation}
where f is the function of FPR and t is of TPR, the trajectory is partitioned in n-1 sections.
\subsection{PRC and the area under the precision-recall curve (AUPRC)}
\label{sec:3.2}
For the classification problem, the performance is typically defined with the confusion matrix generated by the associated classifier. It is possible to compute the recall (sensitivity), precision and specificity. Similar to ROC curves described in the previous chapter, we connect all the pair of measurements at each threshold to make the PRC plot. Precision-Recall curve plots Precision which is the fraction of observations with a positive predicted value that are truly positive versus Recall, measuring the fraction of the examples with positive labels that get a positive predicted result. \\
The perfect classifier will have a PRC that passes through the upper right corner (corresponding to 100\% precision and 100\% recall). In general, the closer a point is to that position, the better the test is. Figure 1 and 2 show PRC and ROC curve for a perfect test respectively. To compare two classifiers based on PRC, the reasonable measure is the area under precision-recall curve (AUPRC). The higher it is, the better the performance of the classifier is. If the area under ROC (AUC) is less than 0.5, representing the baseline value of random guessing, we should convert the negative predictions into positive ones to get a better result. For AUPRC, the corresponding baseline value is the ratio of positive cases in the distribution expressed as
\begin{equation}
Baseline \, Value =\frac{positive\,cases}{positive\,cases+negative\, cases} .\tag{9}
\end{equation}
\begin{figure*}
  \includegraphics[width=0.75\textwidth]{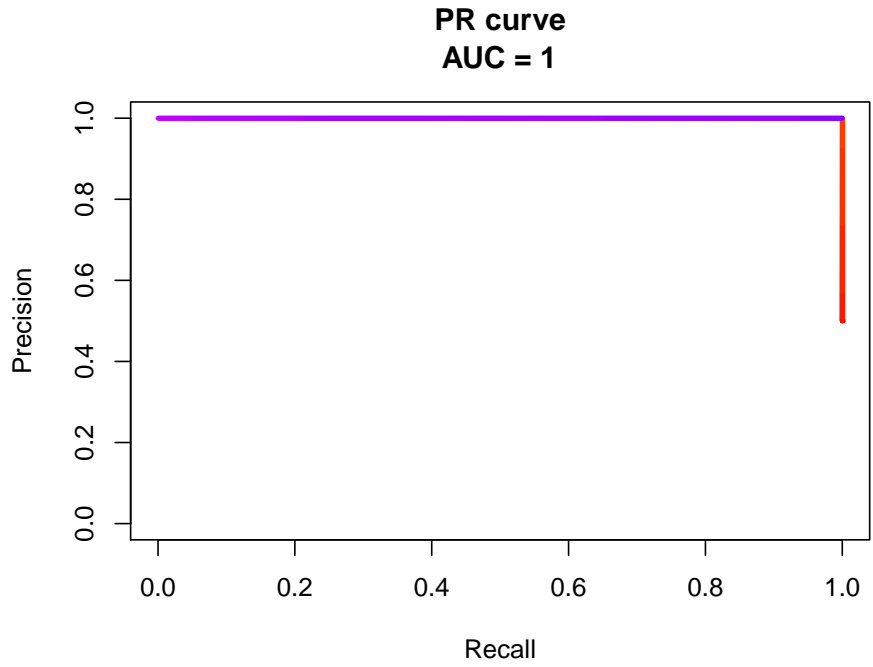}
\caption{PRC for a perfect test}
\label{fig:1}       
\end{figure*}
\begin{figure*}
  \includegraphics[width=0.75\textwidth]{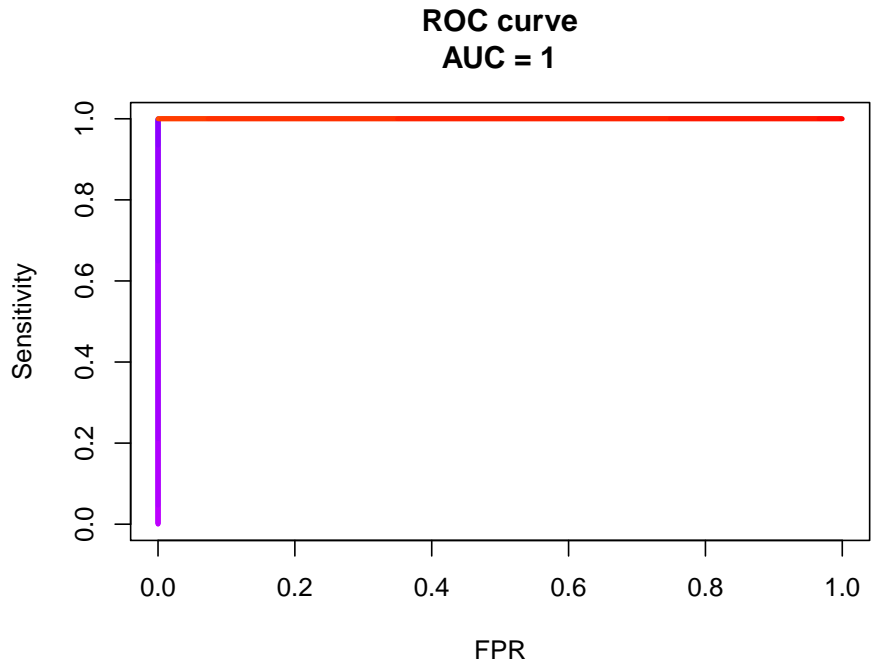}
\caption{ROC curve for a perfect test}
\label{fig:2}       
\end{figure*}
Algorithm 1 below shows the process of calculating the AUPRC and the computation\cite{Ref36} in the case of it less than the baseline value. For the estimation of AUPRC, we also applied the trapezoidal approach: 
\begin{equation}
AUPRC_{trapezoid}=\frac{1}{2}\cdot\sum_{i=1}^{n}(r_{i+1}-r_{i})\cdot(p_{i+1}+p_{i})\tag{10}
\end{equation}
where r is the function of recall and p is of precision, the trajectory is partitioned in n-1 sections. 
\begin{algorithm}[H]
\begin{algorithmic}[1]
\REQUIRE Training data $(x_1,y_1 )$,\ldots,$(x_n,y_n )$, Feature set $f_i \in F$, target vector $y_i \in \{-1,+1\}$. 
\ENSURE Area under precision-recall curve (AUPRC) of feature $f_i$.
\STATE \textbf{sort} data $(X[f_i ], Y)$ by feature
\STATE \textbf{set} $uniq\_ values_{f_i} \leftarrow sort (unique(f_i))$
\STATE \textbf{set} $total\_ positives \leftarrow length (which(Y==1))$
\STATE \textbf{set} $total\_ negatives \leftarrow length (which(Y==-1))$
\STATE \textbf{set} $PRC_{baseline} \leftarrow \frac{total\_ positives}{total\_ positives+total\_ negatives}$
\STATE \textbf{set} $recall\_ array \leftarrow rep(0,length(uniq\_ values_{f_i})$
\STATE \textbf{set} $precision\_ array \leftarrow rep(0,length(uniq\_ values_{f_i})$
\STATE \textbf{set} $AUPRC_{f_i} \leftarrow 0$
\FOR{$j=1$ to $length(uniq\_ values_{f_i})$}
\STATE $indice \leftarrow  which(X[f_i]\leq uniq\_ values_{f_i}[j])$
\STATE $recall\_ array \leftarrow \frac{length(which((X[f_i ],Y)[indice,"Y"]==1))}{total\_ positives}$
\STATE $precision\_ array \leftarrow \frac{length(which((X[f_i ],Y)[indice,"Y"]==1))}{length(indice)}$
\IF{$precision\_ array[j]< PRC_{baseline}$}
\STATE $recall\_ array[j] \leftarrow 1-recall\_ array[j]$
\STATE $precision\_ array[j] \leftarrow \frac{total\_ positives-length(which((X[f_i],Y)[indice,"Y"]==1))}{nrow(X[f_i],Y  )-length(indice)}$
\ENDIF
\IF{$j==1$}
\STATE  $AUPRC_{f_i} \leftarrow \frac{recall\_ array[j]\cdot(1+precision\_ array[j])}{2}$
\ELSE \STATE $AUPRC_{f_i} \leftarrow AUPRC_{f_i}+$\STATE$\frac{(recall\_ array[j]-recall\_ array[j-1])\cdot(precision\_ array[j]+precision\_ array[j-1])}{2}$
\ENDIF
\ENDFOR
\RETURN $AUPRC_{f_i}$
\STATE \textbf{end}
\end{algorithmic}
\caption{AUPRC\_calculation}
\label{al1}
\end{algorithm}
      The Precision-recall curve is usually a tortuous curve, fluctuating up and down. Consequently, PRC tends to intersect each other much more often than the ROC curve. The main difference between them is that the number of true-negative results is not used for constructing the PRC. Ekelund\cite{Ref37} illustrates that the precision-recall curves are not impacted by the addition of patients without disease which are the true-negative cases. Evaluation of classifiers via receiver operating characteristics (ROC) curve have been commonly used. However, for imbalanced data, the ROC may tend to give an optimistic view since the classifier is more likely to assign the observation as the majority class. In many real-life problems, we care more about the true positive cases, e.g., fraud detection and disease diagnosis. The Precision-Recall plot is more informative than the ROC Plot when evaluating binary classifiers on imbalanced datasets\cite{Ref38}. Therefore, we propose the classification trees based on maximizing the area under the Precision-Recall curve.
\subsection{Feature selection}
\label{sec:3.3}
Feature selection is deciding which variable to include in the model. The data features that are used to train tree-based models have a huge influence on the performance we can achieve. In the case of the PRC classification tree, it is done by Algorithm 2. It lets AUPRC decide which feature variable is useful to include in every split of tree node. Typically, the feature that yields the largest AUPRC value should be selected. In Algorithm 2, the function “AUPRC\_calculation” is implemented by Algorithm 1. The below pseudo code illustrates the process of generating the suitable feature with the largest AUPRC value.
\begin{algorithm}[H]
\begin{algorithmic}[1]
\REQUIRE Training data $(x_1,y_1 )$,\ldots,$(x_n,y_n )$, Feature set $f_i \in F$, target vector $y_i \in \{-1,+1\}$.
\ENSURE Feature $f_i$ with the largest AUPRC.
\\ \textbf{Require}: AUPRC\_calculation. 
\STATE \textbf{set} $Max_f  \leftarrow NULL$
\STATE \textbf{set} $Max_{AUPRC} \leftarrow 0$
\FOR{each feature $f_i \in F$}
\STATE $Temp_{AUPRC} \leftarrow AUPRC\_calculation(X[f_i],Y  )$
\IF{$Temp_{AUPRC}> Max_{AUPRC}$}
\STATE $Max_f \leftarrow f_i$
\STATE $Max_{AUPRC} \leftarrow Temp_{AUPRC}$
\ENDIF
\ENDFOR
\RETURN $Max_f, Max_{AUPRC}$
\STATE \textbf{end}
\end{algorithmic}
\caption{Feature\_Selection\_by\_AUPRC}
\label{al2}
\end{algorithm}
\subsection{Threshold selection}
\label{sec:3.4}
It is important to calculate the optimum threshold for each split in the PRC Tree. This could be done by selecting the largest F1-score in the chosen feature variable. The F-score is a measure of accuracy which is the harmonic mean of precision and recall.\\
          Note that the F-score does not take the true negatives into account. That’s the reason that it is more effective than other measures in the classification for imbalanced data. Algorithm 3 illustrates the procedure of threshold selection. The basic goal is to choose the cutoff point with the largest F1 after comparing all the scores for the threshold set of the feature achieved by Algorithm 2.
\begin{algorithm}[H]
\begin{algorithmic}[1]
\REQUIRE PRC matrix ($recall\_ array,precison\_ array$) of selected feature$f_i$, where $f_i \in F$; Threshold set $\Theta$.
\ENSURE Selected splitting threshold $\theta$.
\\ \textbf{Require}: function $HarmonicMean$. 
\STATE \textbf{set} $Max_{F1}  \leftarrow 0$
\STATE \textbf{set} $Max_{\theta}  \leftarrow NULL$
\STATE \textbf{set} $uniq\_ split_{f_j} \leftarrow sort(unique(f_i))$
\FOR{{$j=1$ to $length(uniq\_ split_{f_j})$}}
\STATE $Temp_{F1} \leftarrow HarmonicMean(recall\_ array[j],precison\_ array[j])$
\IF{$Temp_{F1}> Max_{F1}$}
\STATE $Max_{F1} \leftarrow Temp_{F1}$
\STATE $Max_{\theta} \leftarrow uniq\_ split_{f_j}[j]$
\ENDIF
\ENDFOR
\RETURN $Max_{\theta}$
\STATE \textbf{end}
\end{algorithmic}
\caption{Threshold\_Selection\_by\_F1-score}
\label{al3}
\end{algorithm}
\subsection{PRC Tree algorithm}
\label{sec:3.5}
         The following Algorithm 4 is employed to generate the PRC Tree. The previous Algorithm 2 and 3 complete a node split process in each stage. Instead of measuring by traditional Gini impurity or information gain, the features and threshold are selected by AUPRC and F1-score respectively. Algorithm 4 “PRC\_Tree” integrates the previous algorithms to generate each split node until the corresponding stopping criteria is met. Each node has scores for different classes, measuring the percentage of each class in it. It is called nodescore in the algorithm. The corresponding nodelabel is the majority class of this node. By doing so, it could be easy to achieve the majority target label in the Terminal node. Below is the pseudo code to build PRC classification tree recursively. The prediction of a PRC Tree, $\mathcal{T}$, with $K$ terminal nodes and depth $L$, can  be written as
\begin{equation}
g(x_i;\hat{y},K,L)= \sum_{i=1}^{K}\hat{y_{k}}\mathbbm{1}_{\{x_i\in C_k(L)\}},\tag{11}
\end{equation} 
where $C_k(L)$ is one of the $K$ partitions of the data. Each partition is a product of up to $L$ indicator function of the features which are selected by AUPRC algorithm.
\begin{algorithm}[H]
\begin{algorithmic}[1]
\REQUIRE Training data $(x_1,y_1 )$,\ldots,$(x_n,y_n )$, Feature set $f_i \in F$, target vector $y_i \in \{-1,+1\}$; stopping criterion (maximum tree depth, minimum leaf size); $N_f$, the number of features used in each split.
\ENSURE PRC tree.
\\ \textbf{Require}: Feature\_Selection\_by\_AUPRC and Threshold\_Selection\_by\_F1-score
\STATE \textbf{Do} the nodescore and nodelabel for the root node.
\IF{the stopping criterion is met}
\STATE \textbf{return} PRC tree
\ELSE 
\STATE sample $N_f$ features from the feature set $F$
\STATE \textbf{set} the selected features as $F^{'}$
\STATE $(Max_f, Max_{AUPRC}) \leftarrow Feature\_Selection\_by\_AUPRC(X,Y,F^{'})$
\STATE $Max_{\theta} \leftarrow Threshold\_Selection\_by\_F1-score(Max_f, Max_{AUPRC})$
\STATE $\{(X,Y)_{left},(X,Y)_{right}\} \leftarrow Node\_Split(Max_f, Max_{\theta})$
\STATE apply the function $PRC\_Tree\,(maximum tree depth\leftarrow maximum tree depth-1)$ to the subsets $\{(X,Y)_{left},(X,Y)_{right}\}$ recursively until resulting nodes are pure or the stopping criteria is met 
\ENDIF
\RETURN PRC tree
\STATE \textbf{end}
\end{algorithmic}
\caption{PRC\_Tree}
\label{al4}
\end{algorithm}

\subsection{PRC random forest algorithm}
\label{sec:3.6}
Random forests are an ensemble learning method for classification by constructing a multiple of classification trees. We can build our PRC random forest by treating the PRC tree as the base classifier. The PRC random forest has competitive predictive performance and provides a reliable feature importance estimate. Algorithm 5 “PRC\_Random\_Forest” states how to build the forests based on the PRC tree. The parameter $N_t$ is used to decide the number of trees to form the “forest”. As discussed before, the number of features $N_f$ for each node split is randomly chosen from the feature set F. It can decrease the prediction error of the model by doing so. Below is the procedure of our “forests”.

\begin{algorithm}[H]
\begin{algorithmic}[1]
\REQUIRE Training data $(x_1,y_1 )$,\ldots,$(x_n,y_n )$, Feature set $f_i \in F$, target vector $y_i \in \{-1,+1\}$; Number of trees $N_t$; Number of features for each node split $N_f$.
\ENSURE PRC random forest $\Re$.
\\ \textbf{Require}: PRC\_Tree
\STATE \textbf{Set} $\Re \leftarrow NULL$.
\FOR{{$j=1$ to $N_t$}}
\STATE generate bootstrap sample $(X,Y)^j$
\STATE for each node split, generate $F^{'}$ by randomly choosing $N_f$ features from $F$
\STATE $prc\_tree_j \leftarrow PRC\_Tree((X,Y)^j,F^{'})$
\STATE append $prc\_tree_j\, to\, \Re$
\ENDFOR
\RETURN PRC random forest $\Re$
\STATE \textbf{end}
\end{algorithmic}
\caption{PRC\_Random\_Forest}
\label{al5}
\end{algorithm}

\subsection{PRC-ROC Classification Trees }
\label{sec:3.7}
     Each evaluation approach has its own merits and demerits. As an aside, it is comprehensive to combine the PRC and ROC classification trees in meaningful and optimal ways. The following chapters will explore the method based on PRC and ROC curves in more detail.
\subsubsection{Feature selection}
\label{sec:3.7.1}
The one valuable feature of the RF is that we can use the out-of-bag (OOB) samples to make objective performance judgements based on the training data only. The OOB error is an error estimation technique often used to evaluate the accuracy of a random forest. We will use the weighted average of both areas to do the feature selection. Based on the training data and OOB accuracy score, we can tune the parameter to find the optimal weighted parameter for a given data. The parameter can be expressed as 
\begin{equation}
a=\frac{(1-OOB\_error_{PRC RF})}{(1-OOB\_error_{PRC RF}+1-OOB\_error_{ROC RF} )}.\tag{12}
\end{equation}
Hence, the weighted average of both areas becomes
\begin{equation}
Area_{weighted}=a(AUPRC) + (1-a)(AUC), where\, 0\leq a\leq1.\tag{13}
\end{equation}
In the case of the PRC-ROC classification tree, it is done by Algorithm 7. Instead of AUPRC, it lets the weighted average of AUPRC and AUC decide which feature variable is useful to include in every split of tree node. The feature that yields the largest mean value should be selected in a similar fashion. Algorithm 6\cite{Ref25} shows how the function “AUC\_calculation” is implemented. Then we achieves the algorithm by integrating it to the previous feature selection Algorithm 2.
\begin{algorithm}[H]
\begin{algorithmic}[1]
\REQUIRE Training data $(x_1,y_1 )$,\ldots,$(x_n,y_n )$, Feature set $f_i \in F$, target vector $y_i \in \{-1,+1\}$. 
\ENSURE Area under curve (AUC) of feature $f_i$.
\STATE \textbf{sort} data $(X[f_i ], Y)$ by feature
\STATE \textbf{set} $uniq\_ values_{f_i} \leftarrow sort (unique(f_i))$
\STATE \textbf{set} $total\_ positives \leftarrow length (which(Y==1))$
\STATE \textbf{set} $total\_ negatives \leftarrow length (which(Y==-1))$
\STATE \textbf{set} $tp\_ array \leftarrow rep(0,length(uniq\_ values_{f_i})$
\STATE \textbf{set} $fp\_ array \leftarrow rep(0,length(uniq\_ values_{f_i})$
\STATE \textbf{set} $AUC_{f_i} \leftarrow 0$
\FOR{$j=1$ to $length(uniq\_ values_{f_i})$}
\STATE $indice \leftarrow  which(X[f_i]\leq uniq\_ values_{f_i}[j])$
\STATE $tp\_ array \leftarrow \frac{length(which((X[f_i ],Y)[indice,"Y"]==1))}{total\_ positives}$
\STATE $fp\_ array \leftarrow \frac{length(which((X[f_i ],Y)[indice,"Y"]==-1))}{total\_ negatives)}$
\IF{$j==1$}
\STATE  $AUC_{f_i} \leftarrow \frac{tp\_ array[j]*fp\_ array[j]}{2}$
\ELSE \STATE $AUC_{f_i} \leftarrow AUC_{f_i}+$
\STATE$\frac{(fp\_ array[j]-fp\_ array[j-1])\cdot(tp\_ array[j]+tp\_ array[j-1])}{2}$
\ENDIF
\ENDFOR
\IF{$AUC_{f_i}< 0.5$}
\STATE $AUC_{f_i} \leftarrow 1-AUC_{f_i}$
\ENDIF
\RETURN $AUC_{f_i}$
\STATE \textbf{end}
\end{algorithmic}
\caption{AUC\_calculation}
\label{al6}
\end{algorithm}

\begin{algorithm}[H]
\begin{algorithmic}[1]
\REQUIRE Training data $(x_1,y_1 )$,\ldots,$(x_n,y_n )$, Feature set $f_i \in F$, target vector $y_i \in \{-1,+1\}$.
\ENSURE Feature $f_i$ with the largest weighted average of AUPRC and AUC.
\\ \textbf{Require}: AUPRC\_calculation, AUC\_calculation. 
\STATE \textbf{set} $Max_f  \leftarrow NULL$
\STATE \textbf{set} $Max_{AUPRC\_AUC} \leftarrow 0$
\FOR{each feature $f_i \in F$}
\STATE $Temp_{AUPRC\_AUC} \leftarrow Area_{weighted}( AUPRC\_calculation(X[f_i],Y),$ \STATE $AUC\_calculation(X[f_i],Y))$
\IF{$Temp_{AUPRC\_AUC}> Max_{AUPRC\_AUC}$}
\STATE $Max_f \leftarrow f_i$
\STATE $Max_{AUPRC\_AUC} \leftarrow Temp_{AUPRC\_AUC}$
\ENDIF
\ENDFOR
\RETURN $Max_f, Max_{AUPRC\_AUC}$
\STATE \textbf{end}
\end{algorithmic}
\caption{Feature\_Selection\_by\_AUPRC\_AUC}
\label{al7}
\end{algorithm}

\subsubsection{Threshold selection}
\label{sec:3.7.2}
To calculate the optimum threshold for each split in the PRC-ROC Tree, we select the largest harmonic mean of three performance metrics, i.e., recall, precision and specificity in the chosen feature variable. The F-score in this case becomes the harmonic mean of precision, recall and specificity:
      \begin{equation}
F_{3}=\frac{3}{recall^{-1}+precision^{-1}+specificity^{-1}} .\tag{14}
\end{equation}
          Algorithm 8 illustrates the procedure of threshold selection. The main goal is to choose the cutoff point with the largest harmonic mean after comparing all the scores for the threshold set of the feature achieved by feature selection.
\begin{algorithm}[H]
\begin{algorithmic}[1]
\REQUIRE PRC\_AUC matrix ($recall\_ array, precison\_ array, fp\_ array$) of selected feature$f_i$, where $f_i \in F$; Threshold set $\Theta$.
\ENSURE Selected splitting threshold $\theta$.
\\ \textbf{Require}: function $HarmonicMean$. 
\STATE \textbf{set} $Max_{F3}  \leftarrow 0$
\STATE \textbf{set} $Max_{\theta}  \leftarrow NULL$
\STATE \textbf{set} $uniq\_ split_{f_j} \leftarrow sort(unique(f_i))$
\FOR{{$j=1$ to $length(uniq\_ split_{f_j})$}}
\STATE $Temp_{F3} \leftarrow HarmonicMean(recall\_ array[j],precison\_ array[j],fp\_ array[j])$
\IF{$Temp_{F3}> Max_{F3}$}
\STATE $Max_{F3} \leftarrow Temp_{F3}$
\STATE $Max_{\theta} \leftarrow uniq\_ split_{f_j}[j]$
\ENDIF
\ENDFOR
\RETURN $Max_{\theta}$
\STATE \textbf{end}
\end{algorithmic}
\caption{Threshold\_Selection\_by\_F3-score}
\label{al8}
\end{algorithm}

\subsubsection{PRC-ROC Tree algorithm}
\label{sec:3.7.3}
         The following Algorithm 9 is employed to generate the PRC-ROC Tree. The following algorithm “PRC-ROC\_Tree” integrates the previous algorithms to generate each split node until the corresponding stopping criteria is met. Each node has scores for different classes, measuring the percentage of each class in it. It is called nodescore in the algorithm. The corresponding nodelabel is the majority class of this node. By doing so, it could be easy to achieve the majority target label in the Terminal node. Below is the pseudo code to build the classification tree recursively. The prediction of a PRC-ROC Tree, $\mathcal{T}$, with $K$ terminal nodes and depth $L$, can also be written as
\begin{equation}
g(x_i;\hat{y},K,L)= \sum_{i=1}^{K}\hat{y_{k}}\mathbbm{1}_{\{x_i\in C_k(L)\}},\tag{15}
\end{equation} 
where $C_k(L)$ is one of the $K$ partitions of the data. Each partition is a product of up to $L$ indicator function of the features which are selected by AUPRC\_AUC algorithm.
\begin{algorithm}[H]
\begin{algorithmic}[1]
\REQUIRE Training data $(x_1,y_1 )$,\ldots,$(x_n,y_n )$, Feature set $f_i \in F$, target vector $y_i \in \{-1,+1\}$; stopping criterion (maximum tree depth, minimum leaf size); $N_f$, the number of features used in each split.
\ENSURE PRC-ROC tree.
\\ \textbf{Require}: Feature\_Selection\_by\_AUPRC\_AUC and Threshold\_Selection\_by\_F3-score
\STATE \textbf{Do} the nodescore and nodelabel for the root node.
\IF{the stopping criterion is met}
\STATE \textbf{return} PRC-ROC tree
\ELSE 
\STATE sample $N_f$ features from the feature set $F$
\STATE \textbf{set} the selected features as $F^{'}$
\STATE $(Max_f, Max_{AUPRC\_AUC}) \leftarrow Feature\_Selection\_by\_AUPRC\_AUC(X,Y.F^{'})$
\STATE $Max_{\theta} \leftarrow Threshold\_Selection\_by\_F3-score(Max_f, Max_{AUPRC\_AUC})$
\STATE $\{(X,Y)_{left},(X,Y)_{right}\} \leftarrow Node\_Split(Max_f, Max_{\theta})$
\STATE apply the function $PRC-ROC\_Tree\,(maximum tree depth\leftarrow maximum tree depth-1)$ to the subsets $\{(X,Y)_{left},(X,Y)_{right}\}$ recursively until resulting nodes are pure or the stopping criteria is met 
\ENDIF
\RETURN PRC-ROC tree
\STATE \textbf{end}
\end{algorithmic}
\caption{PRC-ROC\_Tree}
\label{al9}
\end{algorithm}

\subsubsection{PRC-ROC random forest algorithm}
\label{sec:3.7.4}
    We can build our PRC-ROC random forest by treating the PRC-ROC tree as the base classifier in a similar way. The PRC-ROC random forest has competitive predictive performance and provides a reliable feature importance estimate. Algorithm 10 “PRC-ROC\_Random\_Forest” states how to build the forests based on the PRC-ROC tree. The parameter $N_t$ is used to decide the number of trees to form the “forest”. As discussed before, the number of features $N_f$ for each node split is randomly chosen from the feature set $F$. It can decrease the prediction error of the model by doing so.
\begin{algorithm}[H]
\begin{algorithmic}[1]
\REQUIRE Training data $(x_1,y_1 )$,\ldots,$(x_n,y_n )$, Feature set $f_i \in F$, target vector $y_i \in \{-1,+1\}$; Number of trees $N_t$; Number of features for each node split $N_f$.
\ENSURE PRC-ROC random forest $\Re$.
\\ \textbf{Require}: PRC-ROC\_Tree
\STATE \textbf{Set} $\Re \leftarrow NULL$.
\FOR{{$j=1$ to $N_t$}}
\STATE generate bootstrap sample $(X,Y)^j$
\STATE for each node split, generate $F^{'}$ by randomly choosing $N_f$ features from $F$
\STATE $prc\_tree_j \leftarrow PRC-ROC\_Tree((X,Y)^j,F^{'})$
\STATE append $prc\_tree_j\, to\, \Re$
\ENDFOR
\RETURN PRC-ROC random forest $\Re$
\STATE \textbf{end}
\end{algorithmic}
\caption{PRC-ROC\_Random\_Forest}
\label{al10}
\end{algorithm}

\subsection{Complexity of algorithm}
\label{sec:3.8}
Let us consider a dataset with input size $n$. Time complexity is the measurement of the speed of the algorithm when performing for the input data. Random Forest is the ensemble model of decision trees. The situation is same to our proposed algorithms. For each tree, it will take $O(N_f\ast nlog(n))$, where $N_f$ is the number of features for each node split. For the random forest algorithm, the complexity would be $O(N_t\ast N_f\ast nlog(n))$ and $N_t$ is the number of the trees.

\section{Experimental Studies }
\label{sec:4}
        The experimental studies fall into two parts. Experiments in the first part are designed for simulation study. The goal is to evaluate the performance on classifying the imbalanced simulated data of binary classes. Experiments in the second part focus on several real-world data sets.\\
        The following step after implementing our proposed algorithm is to find out how effective the model is based on the metrics and datasets. Five performance metrics as follows are used to make evaluation between different algorithms in our analysis. Accuracy is the number of correct predictions made by the model over all kinds of predictions made. Specificity is a measure that tells us what proportion of the majority class, are predicted correctly. Precision shows us what proportion of the predicted minority class which is actually positive. Recall or sensitivity tells us the proportion of cases that actually belong to the minority class is identified by the algorithm as positive cases. F1-score, measuring both precision and recall, is the harmonic mean of them.
\subsection{Simulation study}
\label{sec:4.1}

          In the simulation experimental setting, the number of simulated data sets for classifying binary cases is set to 5,000 each. The feature variables for each class are generated from multivariate Gaussian distribution with specific mean and standard deviation. For our simulation, we divide the class-imbalance extent into 3 categories, namely “Mild”, “Moderate” and “Extreme”, according to the proportion of minority class. In general, the data set whose minority proportion is between 20\% and 40\% should be placed in the mild category. The criterion for moderate category is from 1\% to 20\% and the extreme case is when the proportion is less than 1\%. Similarly, the dimension of features set to be 5 is considered as low feature category and 20 for the higher dimension. Furthermore, the noisy data may also make it difficult to learn the rare class. In the real world, data contains various types of errors, either random or systematic. Random errors are often referred to as noise. To complete the simulation setting, we also give the scenarios with noisy features for comprehensive comparison. Overall, we have two categories (i.e., easy-to-classify and hard-to-classify) in the simulated data and the following sections will give more detail about setting.\\
    For the simulation study, we compared the performance of our proposed algorithms with that of CART, weighted CART and ROC Tree. The CART and weighted CART are implemented by functions in R package “rpart”\cite{Ref39}. Furthermore, we also compared the performance with random forest algorithms such as standard random forest and weighted version. Random forest (RF) is fit by the function in package “randomForest”\cite{Ref40} and weighted Random Forest (Weighted RF) is implemented in package “wsrf”\cite{Ref13}.
\subsubsection{Easy to classify, mild imbalanced data set with low feature dimension}
\label{sec:4.1.1}
  In this scenario, the degree of class-imbalance is set to be mild, i.e., 30\% of the simulated data points belong to the minority class. Five features are generated randomly from multivariate Gaussian distribution with specifies standard deviation 1. The means of the features for the majority and minority class are set to 0 and 3 respectively. %with $\mathbf{\Bar{X}_{1}}$ and $\mathbf{K_{1}}$ denoting the corresponding mean vector and variance-covariance matrix. 
  Figure 3 shows the distribution of the data with 2 features. Table 2 shows the performance metrics of all compared algorithms and it is divided in two. The first block is for single classification tree and the other is for random forest. The best results are denoted in bold respectively. Both PRC algorithms have better performance in F1 score and accuracy than ROC. The PRC is slightly better than CART. In addition, the PRC-ROC tree achieves the best performance in all aspects in the first block. PRC-ROC RF yields the highest recall, accuracy and F1-score among the random forest algorithms.
\begin{figure*}[h]
\centering
\includegraphics[width=0.75\textwidth]{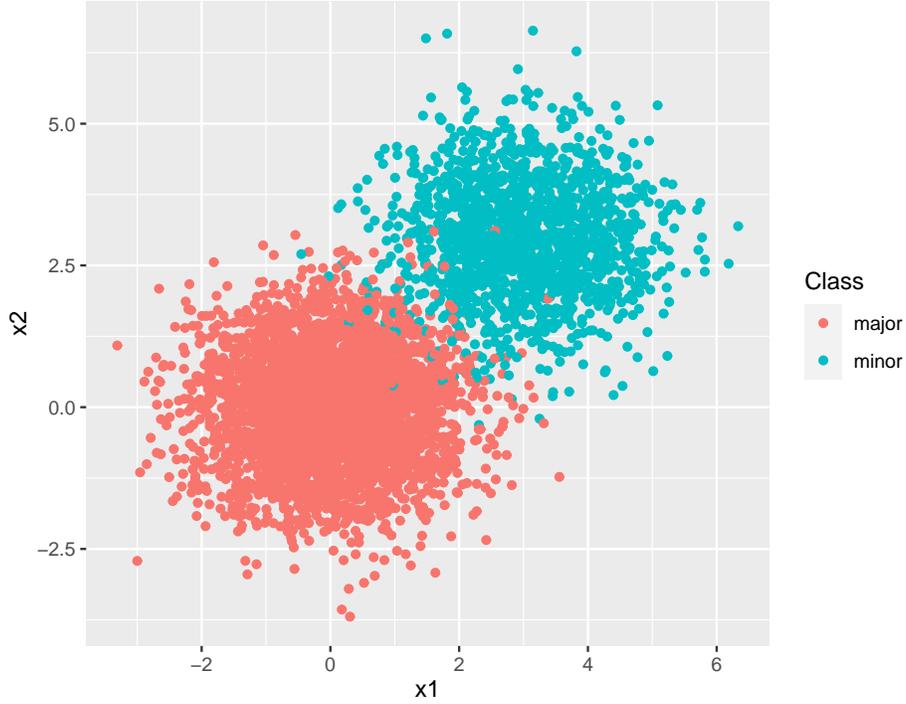}
\caption{Scatterplot of data in scenario 1 (Mild imbalanced extent with low feature dimension) for easy-to-classify setting }
\label{fig:3}       
\end{figure*}

\begin{table}[h]
\centering
\caption{Summary of classification results under scenario 1 (Mild imbalanced extent with low feature dimension)}
\label{tab:2}       
\begin{tabular}{llllll}
\hline\noalign{\smallskip}
Algorithms&Recall&Specificity&Precision&Accuracy&F1 Score  \\
\noalign{\smallskip}\hline\noalign{\smallskip}
CART&	0.9782	&0.9933	&0.9846	&0.9887 &0.9814\\
Weighted CART&	0.9760&	0.9722&	0.9391&	0.9733&	0.9572\\
ROC Tree&	0.9214&	\textbf{0.9942}&0.9860&	0.9720&	0.9526\\
PRC Tree&	0.9891&	0.9904&	0.9784&	0.9900&	0.9837\\
PRC-ROC Tree&\textbf{0.9934}&\textbf{0.9952}&\textbf{0.9891}&\textbf{0.9947}&\textbf{0.9912}\\
\cdashline{1-6}[0.8pt/2pt]
RF&	0.9476	&\textbf{0.9981}&\textbf{0.9954}	&0.9827 &0.9709\\
Weighted RF&	0.9934&	0.9923&	0.9827&	0.9927&	0.9880\\
ROC RF&	0.9476&	0.9923&0.9819&	0.9787&	0.9644\\
PRC RF&	\textbf{1.0000}&	0.9904&	0.9786&	0.9933&	0.9892\\
PRC-ROC RF&\textbf{1.0000}&0.9914&0.9807&\textbf{0.9940}&\textbf{0.9903}\\
\noalign{\smallskip}\hline
\end{tabular}
\end{table}

\subsubsection{Easy to classify, moderate imbalanced data set with higher feature dimension}
In this scenario, the class-imbalance extent becomes moderate. The number of features changes from 5 to 15. Table 3 shows the performance metrics of all the algorithms and the best results are denoted in bold correspondingly. The ROC Tree is perfect in specificity and precision, but the performance of recall is not ideal. We can conclude that the PRC still outperforms both CART in this scenario. Furthermore, PRC-ROC is better than PRC in the five-dimensional evaluation. In the second block, the weighted RF work better than other algorithms under this situation.
\begin{table}[h]
\centering
\caption{Summary of classification results under scenario 2 (Moderate imbalanced extent with higher feature dimension)}
\label{tab:7}       
\begin{tabular}{llllll}
\hline\noalign{\smallskip}
Algorithms&Recall&Specificity&Precision&Accuracy&F1 Score  \\
\noalign{\smallskip}\hline\noalign{\smallskip}
CART&	0.9556	&0.9963	&0.9627	&0.9927   &0.9591\\
Weighted CART&	0.9259&	0.9905&	0.9058&	0.9847&	0.9157\\
ROC Tree&	0.4667&	\textbf{1.0000}&\textbf{1.0000}&0.9520&	0.6364\\
PRC Tree&0.9704	&0.9985	&0.9850	&0.9960  &0.9776\\
PRC-ROC Tree&\textbf{0.9778}&0.9993&0.9925&	\textbf{0.9973}&\textbf{0.9851}\\
\cdashline{1-6}[0.8pt/2pt]
RF&0.9556&0.9832&	0.8487&	0.9807&0.8990\\
Weighted RF&0.9259&\textbf{1.0000}&\textbf{1.0000}&\textbf{0.9933}&\textbf{0.9615}\\
ROC RF&	\textbf{1.0000}&0.9875&	0.8882&	0.9887&0.9408\\
PRC RF&	\textbf{1.0000}&0.9685&	0.7584&	0.9713&0.8626\\
PRC-ROC RF&\textbf{1.0000}&0.9817&	0.8437&	0.9833&0.9152\\
\noalign{\smallskip}\hline
\end{tabular}
\end{table}

\subsubsection{Easy to classify, extreme imbalanced data set with low feature dimension}
Under this scenario, only 1\% of the simulated data points belong to the minority class. The parameter setting of features remains the same as scenario 1. Table 4 shows the performance metrics. The PRC achieves a higher number in recall than the weighted one. Both PRC methods apparently outclass than CART and ROC especially in recall. PRC-ROC obtains larger values in specificity, precision and F1-score than other methods. Although the weighted RF achieves the best specificity and precision, PRC-ROC RF yields the highest recall, accuracy and F1-score among the random forest algorithms.
\begin{table}[h]
\centering
\caption{Summary of classification results under scenario 3 (Extreme imbalanced extent with low feature dimension)}
\label{tab:10}       
\begin{tabular}{llllll}
\hline\noalign{\smallskip}
Algorithms&Recall&Specificity&Precision&Accuracy&F1 Score  \\
\noalign{\smallskip}\hline\noalign{\smallskip}
CART&	0.5833	&0.9980	&0.7000	&0.9947& 0.6363\\
Weighted CART&0.8333&0.9919&0.4545&	0.9907&	0.5882\\
ROC Tree&	0.3333&	0.9980&0.5714&0.9927&	0.4210\\
PRC Tree&\textbf{0.9167}&0.9973&	0.7333&	0.9967&	0.8148\\
PRC-ROC Tree&0.7500&\textbf{0.9993}&\textbf{0.9000}&\textbf{0.9973}&\textbf{0.8182}\\
\cdashline{1-6}[0.8pt/2pt]
RF&	0.7500&0.9993&0.9000&\textbf{0.9973}&0.8182\\
Weighted RF&0.5833&\textbf{1.0000}&\textbf{1.0000}&0.9967&0.7368\\
ROC RF&	0.8333&0.9973&0.7143&0.9960&0.7692\\
PRC RF&	\textbf{1.0000}&0.9966&	0.7059&	0.9967&0.8276\\
PRC-ROC RF&\textbf{1.0000}&0.9973&	0.7500&	\textbf{0.9973}&\textbf{0.8571}\\
\noalign{\smallskip}\hline
\end{tabular}
\end{table}

\subsubsection{Hard to classify, mild imbalanced data set with higher feature dimension and noise} 
In this hard-to-classify scenario, the degree of class-imbalance is set to be mild, i.e., 30\% of the simulated data points belong to the minority class. Five features are generated randomly from multivariate Gaussian distribution with specifies standard deviation 1. The means of the features for the majority and minority class are set to 0 and -1 respectively. Figure 4 shows the distribution of the data with 2 features. We can see many overlaps compared to the previous setting.
The setting for this scenario is with ten normal features and five noise features. We change their means to 1 to treat them as noisy features. Table 5 summarizes the performance metrics of the algorithms mentioned and the best performance is denoted in the same way. The results obviously indicate that in this specific case PRC-ROC surpasses other tree-algorithms in every metrics except the recall value. For the comparison of the random forest methods, PRC-ROC RF outperforms others.
\begin{figure*}[h]
\centering
  \includegraphics[width=0.75\textwidth]{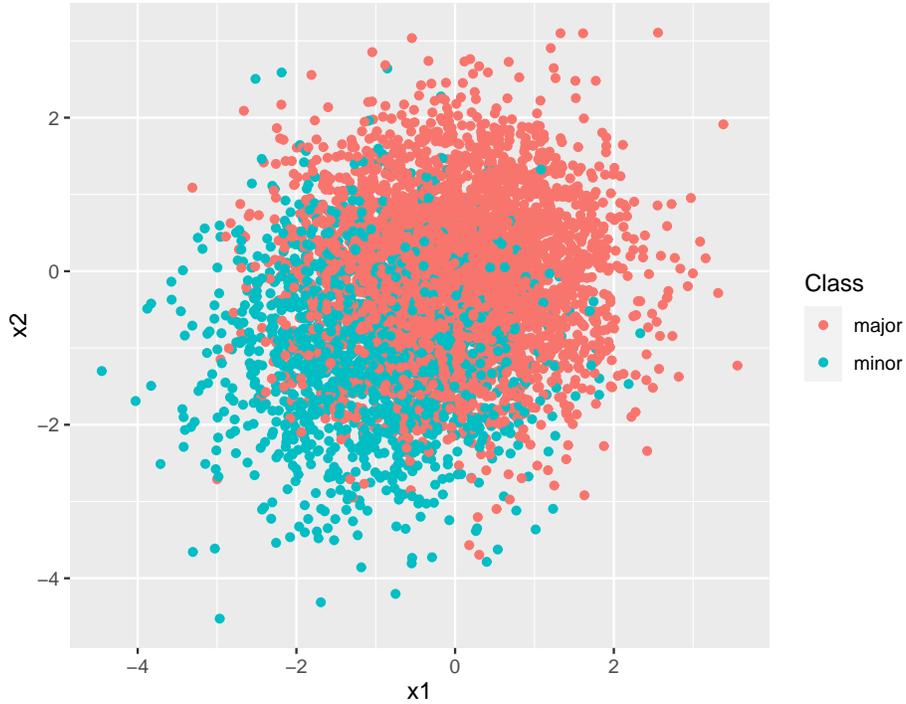}
\caption{Scatterplot of data in scenario 4 (Mild imbalanced extent with low feature dimension) for hard-to-classify setting}
\label{fig:4}       
\end{figure*}
\begin{table}[ht]
\centering
\caption{Summary of classification results under scenario 4 (Mild imbalanced extent with higher feature dimension and noise)}
\label{tab:17}       
\begin{tabular}{llllll}
\hline\noalign{\smallskip}
Algorithms&Recall&Specificity&Precision&Accuracy&F1 Score  \\
\noalign{\smallskip}\hline\noalign{\smallskip}
CART&	0.6372&	0.9141&0.7554&	0.8327 &0.6913\\
Weighted CART&\textbf{0.7846}&	0.8036&0.6245&	0.7980&	0.6955\\
ROC Tree&	0.6689&	0.8687&0.6797&	0.8100&	0.6743\\
PRC Tree&	0.7347&	0.8820&0.7216&	0.8387&	0.7281\\
PRC-ROC Tree&0.7143&\textbf{0.9056}&\textbf{0.7590}&\textbf{0.8493}&\textbf{0.7360}\\
\cdashline{1-6}[0.8pt/2pt]
RF&	0.5011&0.8971&0.6697&0.7807&0.5733\\
Weighted RF&0.7279&	0.9452&0.8470&	0.8813&	0.7829\\
ROC RF&	0.9297&	0.8291&0.6937&	0.8587&	0.7945\\
PRC RF&\textbf{0.9751}	&0.8263&0.7003&0.8700&0.8152\\
PRC-ROC RF&0.9478&\textbf{0.8565}&\textbf{0.7333}&\textbf{0.8833}&\textbf{0.8269}\\
\noalign{\smallskip}\hline
\end{tabular}
\end{table}

\subsubsection{Hard to classify, extreme imbalanced data set with higher feature dimension and noise}
 In this extreme data set, the features’ distribution is the same as scenario 4. Table 6 summarizes the comparison of all mentioned methods and the best results are denoted in bold. The PRC-ROC tree attains the best performance in the last four dimensions and makes huge improvement in F1 score. The ROC RF achieves the highest F1-score in the second block.\\
\begin{table}[h]
\centering
\caption{Summary of classification results under scenario 5 (Extreme imbalanced extent with higher feature dimension and noise)}
\label{tab:25}       
\begin{tabular}{llllll}
\hline\noalign{\smallskip}
Algorithms&Recall&Specificity&Precision&Accuracy&F1 Score  \\
\noalign{\smallskip}\hline\noalign{\smallskip}
CART&	0.2000&	0.9966&	0.2857&	0.9913&	0.2353\\
Weighted CART&	\textbf{0.6000}&	0.9403&	0.0632&	0.9380&	0.1144\\
ROC Tree&	0.1000&	0.9919&0.0769&	0.9860&	0.0869\\
PRC Tree&	0.2000&0.9987&0.5000&	0.9933&0.2857\\
PRC-ROC Tree&0.2000&\textbf{0.9993}&\textbf{0.6667}&\textbf{0.9940}&\textbf{0.3077}\\
\cdashline{1-6}[0.8pt/2pt]
RF&	0.1000&	0.9933&	0.0909&	0.9873&	0.0952\\
Weighted RF&0.1000&	\textbf{0.9960}&\textbf{0.1429}&\textbf{0.9900}&0.1177\\
ROC RF&\textbf{0.7000}	&	0.9322&0.0648&	0.9307&	0.1186\\
PRC RF&	0.6000&0.9483&0.0723&0.9460&0.1290\\
PRC-ROC RF&0.6000&0.9523&0.0779&0.9500&\textbf{0.1381}\\
\noalign{\smallskip}\hline
\end{tabular}
\end{table}

To sum up, the comparisons under most of those cases illustrate that two PRC trees achieve better performance than the both CART trees when the data set is imbalanced in either easy-to-differentiate or hard-to-differentiate case. Especially in the case of the extreme data set, the PRC tree is more powerful to identify the minority class and achieves higher score in precision and recall than traditional CART. The ROC and PRC Tree have different focuses when dealing with imbalanced data.The PRC Tree is better than the ROC Tree when the main goal is to identify the rare cases and achieve the satisfactory performance in F1 score. Moreover,PRC-ROC is slightly better than PRC in our 5-dimensional evaluation in most cases since it inherits the advantage of both classification trees. For the second block, the two PRC RFs also show some advantages in classifying the minority cases. The following section moves on to the hard-to-classify setting. The next section focuses on the evaluation in the real-world data set. 

\subsection{Real-world data set experiment}
\label{sec:4.2}
The PRC and PRC-ROC Tree have shown their advantage in classifying the simulated data set which is severely skewed. In this section we choose several datasets from Kaggle competition and UCI Machine Learning Repository to testify the feasibility of PRC random forests. We will compare the performance with standard classification algorithms such as random forest, SVM and neural network. In addition, weighted CART and ROC random forest are all taken into account. Random forest is fit by the function in package “randomForest”\cite{Ref40}. SVM fits in package “e1071”\cite{Ref41} and the package “neuralnet”\cite{Ref42} is for realizing neural network. Weighted Random Forest (Weighted RF) is implemented in package “wsrf”\cite{Ref13}.\\
The Table 7 summarizes the basic information of the three selected data sets. The first two data sets are about financial distress prediction and default of clients. The remaining one is about breast cancer diagnosis. For the credit data, this study employed a binary variable, default payment (Yes = 1, No = 0), as the dependent variable. Besides, this data set used 23 features as the independent variables. Similarly, the target variable in the second data is denoted by "Financial Distress" if it is greater than -0.50 the company should be considered as healthy (0). Otherwise, it would be regarded as financially distressed (1). We have 83 explanatory variables for it. The binary outcome of interest for the cancer data set that contains 30 attributes is whether a tumor is benign (0) or malignant (1). \\
\begin{table}[h]
\centering
\caption{Summary of selected real-world datasets}
\label{tab:26}    
\resizebox{\columnwidth}{!}{%
\begin{tabular}{lllll}
\hline\noalign{\smallskip}
Dataset Name&	Data Sources&	Number of Observations&	Proportion of Minority Class&	Number of Features  \\
\noalign{\smallskip}\hline\noalign{\smallskip}
Default of Credit Card Clients&	UCI&	30000&	3.7\%&	23\\
Financial Distress Prediction&	Kaggle&	3672&	22\%&	83\\
Breast Cancer Wisconsin&	UCI&	569&	37\%&	30\\
\noalign{\smallskip}\hline
\end{tabular}%
}
\end{table}
    The out-of-bag error is an error estimation technique often used to evaluate the accuracy of a random forest. We split the data into training and testing and use the OOB to estimate the error rate of the training sample. The following table shows the out-of-bag error of all the corresponding random forests and the minimum values are colored in red. Then we still use the separate testing data to further validate our RF model built with the training data and the results are given in the last three tables. \\

\begin{table}[h]
\centering
\caption{Summary of Out-of-bag error rate}
\label{tab:27}    
\resizebox{\columnwidth}{!}{%
\begin{tabular}{lllll}
\hline\noalign{\smallskip}
Dataset Name&	RF&	Weighted RF&	ROC RF&	PRC RF\\
\noalign{\smallskip}\hline\noalign{\smallskip}
Default of Credit Card Clients&	\textcolor{red}{0.1900}&	0.2300	&0.2446&	0.1921\\
Financial Distress Prediction&	0.0414&	0.0500&	0.0345&\textcolor{red}{0.0000}\\
Breast Cancer Wisconsin	&0.1000&0.0912&	0.0714&	\textcolor{red}{0.0455}\\
\noalign{\smallskip}\hline
\end{tabular}%
}
\end{table}

\begin{table}[h]
\centering
\caption{Summary of performance results for Default of Credit Card Clients Data}
\label{tab:28}       
\begin{tabular}{llllll}
\hline\noalign{\smallskip}
Algorithms&Recall&Specificity&Precision&Accuracy&F1 Score\\
\noalign{\smallskip}\hline\noalign{\smallskip}
SVM	&0.3343&	\textbf{0.9628}&	0.7204&	0.8229&	0.4567\\
\textcolor{blue}{Neural Network}&	0.3947&	0.9441&	0.6692&	0.8218&	0.4965\\
\cdashline{1-6}[0.8pt/2pt]
CART&	0.3378	&0.9627	& 0.7218	&0.8236 &0.4602\\
Weighted CART&	0.5788&	0.8253&	0.4870&	0.7704&	0.5289\\
ROC Tree&	\textbf{0.7061}&	0.5330&	0.3022&	0.5716&	0.4233\\
PRC Tree&	0.4187&	0.9200&	0.5997&	0.8083&	0.4931\\
PRC-ROC Tree &	0.3353&	0.9634&	\textbf{0.7241}&	0.8236&	0.4584\\
RF&	0.3493&	0.9608	&0.7187&	0.7839&	0.4701\\
Weighted RF&	0.3558&	0.9570&	0.7032&	0.8231&	0.4968\\
ROC RF&	0.6307&	0.5850&	0.3033&	0.5952&	0.4096\\
\textcolor{green}{PRC RF}&	0.5744&	0.8439&	0.5132&	\textbf{0.8247}&	\textbf{0.5432}\\
PRC-ROC RF &	0.5813&	0.8340&	0.5009&	0.7778&	0.5381\\
\noalign{\smallskip}\hline
\end{tabular}
\end{table}

\begin{table}[h]
\centering
\caption{Summary of performance results for Financial Distress Prediction Data}
\label{tab:29}       
\begin{tabular}{llllll}
\hline\noalign{\smallskip}
Algorithms&Recall&Specificity&Precision&Accuracy&F1 Score\\
\noalign{\smallskip}\hline\noalign{\smallskip}
SVM	&0.0000	&\textbf{1.0000}	&\textbf{NaN}&	0.9600&	\textbf{NaN}\\
\textcolor{blue}{Neural Network}&	0.3100&	0.9774&	0.3514&	0.9528&	0.3294\\
\cdashline{1-6}[0.8pt/2pt]
CART&	0.1951	&0.9830	& 0.3077	&0.9537 &0.2388\\
Weighted CART&	0.8049&	0.8727&	0.1964&	0.8702&	0.3158\\
ROC Tree&	0.0730&	0.9906&	0.2308&	0.9564&	0.1110\\
\textcolor{green}{PRC Tree} &	0.3902&	0.9802&	0.4324&	0.9583&	\textbf{0.4102}\\
PRC-ROC Tree &	0.2439&	0.9934&	0.5882&	\textbf{0.9655}&	0.3448\\
RF&	0.0244&	0.9981&	0.3333	&0.9619	&0.0455\\
Weighted RF&	0.1951&	0.9953&	\textbf{0.6154}&	\textbf{0.9655}&	0.2963\\
ROC RF&	\textbf{0.9512}&	0.8021&	0.1566&	0.8076&	0.2689\\
PRC RF&	0.9024&	0.8633&	0.2033&	0.8648&	0.3318\\
PRC-ROC RF &0.9268&	0.8681&	0.2135&	0.8702&	0.3471\\
\noalign{\smallskip}\hline
\end{tabular}
\end{table}

\begin{table}[h]
\centering
\caption{Summary of performance results for Breast Cancer Data}
\label{tab:30}       
\begin{tabular}{llllll}
\hline\noalign{\smallskip}
Algorithms&Recall&Specificity&Precision&Accuracy&F1 Score\\
\noalign{\smallskip}\hline\noalign{\smallskip}
\textcolor{blue}{SVM}	&0.9672&	0.9909&	0.9833	&0.9616	&\textbf{0.9751}\\
\textcolor{blue}{Neural Network}&	0.9672&	0.9909&	0.9833&	0.9616&	\textbf{0.9751}\\
\cdashline{1-6}[0.8pt/2pt]
CART&	0.9016	&0.9909	& 0.9821	&0.9591 &0.9401\\
Weighted CART&	0.9344&	0.9818&	0.9661&	0.9649&	0.9500\\
ROC Tree&	0.8197&	0.9909&	0.9804&	0.9298&	0.8929\\
PRC Tree &0.9180&	0.9455&	0.9032&	0.9357&	0.9105\\
PRC-ROC Tree &0.9344&	0.9818&	0.9661&	0.9649&	0.9476\\
RF&		0.8689&	0.9909	&0.9814&	0.9473&	0.9217\\
Weighted RF&	0.9180	&\textbf{1.0000}	&\textbf{1.0000}&	0.9707&	0.9572\\
ROC RF&	0.8361&	0.9099&	0.9808	&0.9357	&0.9027\\
PRC RF&	\textbf{0.9836}	&0.9636	&0.9375	&0.9708	&0.9600\\
\textcolor{green}{PRC-ROC RF}&	\textbf{0.9836}&	0.9727&	0.9524&	\textbf{0.9766}&	0.9677\\
\noalign{\smallskip}\hline
\end{tabular}
\end{table}

Then we experiment with real-world data sets and summarize the performance comparison in Tables 9 to 11. The best performance is denoted in bold. Furthermore, we use green to highlight the best tree model and blue to color the best non-tree model. The results are quite consistent with the out-of-bag error rate table. In general, the proposed algorithm can achieve higher recall and F1-score than traditional random forest. PRC random forest achieves the largest accuracy, F1-score and PRC-ROC attains the highest precision in the first case. Despite random forest achieving a slightly higher specificity, the recall and F1-score are much lower than PRC random forest and PRC-ROC RF. This is even more confirmed in the Financial Distress Prediction Data. In this distress data, SVM tends to classify all the cases as the majority class. Weighted RF did a good job in the performance of precision and accuracy. However, PRC random forest and PRC-ROC RF achieve good performance in F1 score on average. To conclude, PRC Random Forest is superior to other algorithms especially traditional random forest in identifying the minority class which is exactly the goal of imbalanced classification. In the last two instances, the PRC-ROC Random Forest has higher accuracy and F1-score compared to other algorithms and PRC RF’s behavior is close to it.

\section{Conclusion}
In this paper, we investigate the PRC classification tree for dealing with class imbalance problem. We also extend this to the general PRC-ROC tree and random forest by setting an additional weight parameter which selects the best weighted combination of the AUPRC and the AUC based on the training data, for each problem at hand. Four new tree-based algorithms (i.e., PRC Tree, PRC RF, PRC-ROC Tree and PRC-ROC Tree RF) are studied along with several existing related algorithms. Table 31 shows the F1-score performance of the proposed classifiers against other tree-based algorithms in all datasets. The results indicate that our methods are superior to other rivals for skewed classification problems.\\
Several research issues are needed for in-depth investigation. Our current PRC tree/forest and PRC-ROC tree/forest are all designed for binary classifications. We shall extend them to accommodate multiple-class classifications arising naturally from many practical scenarios. \\
In our experimental study, we do not consider the interaction term. Hence another related task is to uncover interaction effects in the framework of PRC classification tree. There are absolutely many other potential possibilities about classification of imbalanced data and it will continue to gain increasingly attention in academia and industry.
\bibliographystyle{unsrt}  


\begin{thebibliography}{1}

\bibitem{Ref1}
Ethem Alpaydin, Introduction to Machine Learning. MIT Press (2010)
\bibitem{Ref2}
Wei-Yang Lin, Ya-Han Hu, Chih-Fong Tsai, Machine Learning in Financial Crisis Prediction: A Survey, IEEE Transactions on Systems, Man, and Cybernetics, Part C (Applications and Reviews),42(4), 421-436 (2012)
\bibitem{Ref3}
Jie Sun, Hui  Li, Qinghua  Huang, Kaiyu  He, Predicting financial distress and corporate failure: A review from the state-of-the-art definitions, modeling, sampling, and featuring approaches, Knowledge-Based Systems, 57 (2014)
\bibitem{Ref4}
Leo Breiman, Jerome H Friedman, Richard A Olshen, Charles J Stone, Classification and Regression Trees (1984)
\bibitem{Ref5}
Ho, Tin Kam, Random Decision Forests, Proceedings of the 3rd International Conference on Document Analysis and Recognition, 278–282 (1995)
\bibitem{Ref6}
Leo Breiman, Random Forests, Machine Learning, 45, 5–32 (2001)
\bibitem{Ref7}
Bradley Efron, Bootstrap Methods: Another Look at the Jackknife, The Annals of Statistics, 7(1), 1-26 (1979)
\bibitem{Ref8}
Corinna Cortes, Vladimir Vapnik, Support-vector networks, Machine Learning, 20(3), 273–297 (1995)
\bibitem{Ref9}
R. Akbani, S. Kwek, N. Jakowicz, Applying support vector machines to imbalanced datasets, Proceedings of 15th European Conference on Machine Learning, 39-50 (2004)
\bibitem{Ref10}
K. Carvajal, M. Chacon, D. Mery, G. Acuna, Neural network method for failure detection with skewed class distribution, Insight-Non-Destructive Testing and Condition Monitoring, 46(7), 399-402 (2004)
\bibitem{Ref11}
Charles Elkan, The foundations of cost-sensitive learning. International joint conference on artificial intelligence, 2, 973-978 (2001)
\bibitem{Ref12}
Chao Chen, Andy Liaw, Leo Breiman, Using Random Forest to Learn Imbalanced Data, Research article UC Berkeley, 1-12 (2004)
\bibitem{Ref13}
He Zhao, Graham J.ANU Williams, Joshua Zhexue Huang, Qinghan Meng, Baoxun Xu, Weighted Subspace Random Forest for Classification, International Journal of Data Warehousing and Mining, 8(2), 44-63 (2012)
\bibitem{Ref14}
SungHwan Kim, Weighted K-means support vector machine for cancer prediction, 1162. SpringerPlus (2016)
\bibitem{Ref15}
Bartosz Krawczyk, Michal Wozniak, Gerald Schaefer, Cost-sensitive decision tree ensembles for effective imbalanced classification, Applied Soft Computing, 14, 554-562 (2014)
\bibitem{Ref16}
Saumya Debray, Sampath Kannan, Mukul Paithane, Weighted Decision Trees, Proceedings of the Joint International Conference and Symposium on Logic Programming, 654-668 (1992)
\bibitem{Ref17}
Matjaz Kukar, Igor Kononenko, Cost-Sensitive Learning with Neural Networks, Proceedings of the 13th European Conference on Artificial Intelligence (ECAI-98), 445-449 (1998)
\bibitem{Ref18}
Qiuyan Yan, Shixiong Xia, Fanrong Me, Optimizing Cost-Sensitive SVM for Imbalanced Data: Connecting Cluster to Classification, ArXiv (2017)
\bibitem{Ref19}
Long Wang, Cost-sensitive Boosted ROC Classification Trees, Doctoral dissertation, The Graduate School, Stony Brook University (2019)
\bibitem{Ref20}
Robert A. Sowah, Moses A. Agebure, Godfrey A. Mills, Koudjo M. Koumadi, Seth Y. Fiawoo, New Cluster Undersampling Technique for Class Imbalance Learning, International Journal of Machine Learning and Computing, 6(3) (2016)
\bibitem{Ref21}
N. V. Chawla, K. W. Bowyer, L. O. Hall, W. P. Kegelmeyer, SMOTE: Synthetic Minority Over-sampling Technique, Journal Of Artificial Intelligence Research, 16, 321-357 (2002)
\bibitem{Ref22}
Haibo He, Yang Bai, Edwardo A. Garcia, Shutao Li, ADASYN: Adaptive Synthetic Sampling Approach for Imbalanced Learning, International Joint Conference on Neural Networks (2008)
\bibitem{Ref23}
Bowen Song, ROC Random Forest and Its Application. Doctoral dissertation, The Graduate School, Stony Brook University (2015)
\bibitem{Ref24}
Song B, Zhang G, Zhu W, Liang Z, ROC operating point selection for classification of imbalanced data with application to computer-aided polyp detection in CT colonography, Int J Comput Assist Radiol Surg, 9(1), 79-89 (2014)
\bibitem{Ref25}
Jiaju Yan, Multi-Class ROC Random Forest for Imbalanced Classification. Doctoral dissertation, The Graduate School, Stony Brook University (2017)
\bibitem{Ref26}
Yanmin Sun, Kamel, M. S., Wong, A. K., Wang, Y, Cost-sensitive boosting for classification of imbalanced data, Pattern Recognition, 40(12), 3358-3378 (2007)
\bibitem{Ref27}
Yanmin Sun, Cost-Sensitive Boosting for Classification of Imbalanced Data, Doctoral dissertation, University of Waterloo (2007)
\bibitem{Ref28}
Qiujie Li and Yaobin Mao, A review of boosting methods for imbalanced data classification, Pattern Anal Applic, 17, 679–693 (2014)
\bibitem{Ref29}
Hamed Masnadi-Shirazi, Nuno Vasconcelos, Cost-Sensitive Boosting, IEEE Transactions on pattern analysis and machine intelligence, 33(2), 294-309 (2010)
\bibitem{Ref30}
Nitesh V. Chawla, Aleksandar Lazarevic, Lawrence O. Hall, Kevin Bowyer, SMOTEBoost: Improving Prediction of the Minority Class in Boosting, PKDD, 107-119 (2003)
\bibitem{Ref31}
Wenhao Zhang, Ramin Ramezani, Arash Naeim, WOTBoost: Weighted oversampling technique in boosting for balanced learning, IEEE BigData (2019)
\bibitem{Ref32}
Chris Seiffert, Taghi M. Khoshgoftaar, Jason Van Hulse, Amri Napolitano, RUSBoost: A Hybrid Approach to Alleviating Class Imbalance, IEEE Transactions on Systems, Man, and Cybernetics - Part A: Systems and Humans, 40(1) (2010)  
\bibitem{Ref33}
Farshid Rayhan, Sajid Ahmed, Asif Mahbub, Md. Rafsan Jani, Swakkhar Shatabda, Dewan Md. Farid, Chowdhury Mofizur Rahman, MEBoost: Mixing Estimators with Boosting for Imbalanced Data Classification, Pattern Recognition, 46(12), 3460-3471 (2013)
\bibitem{Ref34}
Hongyu Guo, Herna L Viktor, Learning from Imbalanced Data Sets with Boosting and Data Generation: The DataBoost-IM Approach, ACM SIGKDD Explorations Newsletter, 6(1) (2004)
\bibitem{Ref35}
Mikel Galar, Alberto Fernández, Edurne Barrenechea, Francisco Herrera, EUSBoost: Enhancing ensembles for highly imbalanced data-sets by evolutionary undersampling, Pattern Recognition, 46(12), 3460-3471 (2013)
\bibitem{Ref36}
Kendrick Boyd, Kevin H. Eng, C. David Page, Area Under the Precision-Recall Curve: Point Estimates and Confidence Intervals, Machine Learning and Knowledge Discovery in Databases, 451-466 (2013)
\bibitem{Ref37}
Suzanne Ekelund, Precision-recall curves – what are they and how are they used? Information Management (2017)
\bibitem{Ref38}
Takaya Saito, Marc Rehmsmeier, The Precision-Recall Plot Is More Informative than the ROC Plot When Evaluating Binary Classifiers on Imbalanced Datasets. Journal of PLOS ONE (2015)
\bibitem{Ref39}
Terry Therneau, Beth Atkinson, Brian Ripley, rpart: Recursive Partitioning and Regression Trees (2019)
\bibitem{Ref40}
Andy Liaw, Matthew Wiener, Breiman and Cutler's Random Forests for Classification and Regression (2018)
\bibitem{Ref41}
David Meyer, et al,  Functions for latent class analysis, short time Fourier transform, fuzzy clustering, support vector machines, shortest path computation, bagged clustering, naive Bayes classifier (2019)
\bibitem{Ref42}
Stefan Fritsch, et al, Training of Neural Networks (2019)
% Format for books Author, Article title, Journal, Volume, page numbers (year)

\end{thebibliography}

\end{document}